\documentclass[letterpaper, 10 pt, conference]{ieeeconf}  
\IEEEoverridecommandlockouts      % This command is only neede if 
\usepackage{graphics}   % for pdf, bitmapped graphics files
\usepackage{epsfig}     % for postscript graphics files
\usepackage{times}      % assumes new font selection scheme installed
\usepackage{amsmath}    % assumes amsmath package installedd
\usepackage{amssymb}    % assumes amsmath package installed

\usepackage{xcolor}
\usepackage{orcidlink}
\usepackage{cite}
\usepackage{balance}
\usepackage{booktabs}
\usepackage{algorithm}
\usepackage{algorithmic}
\usepackage{xfrac}
\usepackage{hyperref}
\hypersetup{hidelinks}
\title{\LARGE \bf
A Real-Time Multi-Model Parametric Representation of Point Clouds
}

\author{
Yuan Gao\orcidlink{0009-0008-8598-0613}, Wei Dong\orcidlink{0000-0003-2640-1585}
\thanks{This work was supported by AAAA \emph{(Corresponding author: Wei Dong.)}}
\thanks{All authors are with the School of Mechanical Engineering, Shanghai Jiao Tong University, Shanghai 200240, China (e-mail: GaoY-23@sjtu.edu.cn; dr.dongwei@sjtu.edu.cn).}
}

\begin{document}

\maketitle
\thispagestyle{empty}
\pagestyle{empty}

\begin{abstract}

In recent years, parametric representations of point clouds have been widely applied in tasks such as memory-efficient mapping and multi-robot collaboration. 
Highly adaptive models, like spline surfaces or quadrics, are computationally expensive in detection or fitting. In contrast, real-time methods, such as Gaussian mixture models or planes, have low degrees of freedom, making high accuracy with few primitives difficult.
To tackle this problem, a multi-model parametric representation with real-time surface detection and fitting is proposed.
Specifically, the Gaussian mixture model is first employed to segment the point cloud into multiple clusters. 
Then, flat clusters are selected and merged into planes or curved surfaces. Planes can be easily fitted and delimited by a 2D voxel-based boundary description method. Surfaces with curvature are fitted by B-spline surfaces and the same boundary description method is employed.
Through evaluations on multiple public datasets, the proposed surface detection exhibits greater robustness than the state-of-the-art approach, with 3.78 times improvement in efficiency. Meanwhile, this representation achieves a 2-fold gain in accuracy over Gaussian mixture models, operating at 36.4 fps on a low-power onboard computer.

\end{abstract}

% KEY WORDS
% Mapping, Object detection/segmentation/categorization, Computational Geometry, 

\section{Introduction}

Parametric representation of point clouds plays an important role in robotic perception. For simultaneous localization and mapping (SLAM), point cloud parametric representation can enable the system to apply line, plane or surface features in the environment, thus decreasing the memory consumption and improving efficiency \cite{a0, a1, e1, e2, e3}.
In multi-robot systems, compressing point clouds with parametric models sharply decreases the requirement of bandwidth for exchanging map information \cite{a2,a3,a4}. Moreover, by extracting specific environmental feature information, robots can perform tasks, such as docking and grasping, automatically \cite{a5, a6}.

Due to the limited computational resources of robotic systems, achieving high-precision point cloud parameterization in real-time remains challenging. Recently, parameterization methods based on 3D Gaussian splatting (3DGS) have attracted significant attention \cite{b0,b1,b1_after}. While effective for accurate 3D reconstruction, these methods are unsuitable for real-time robotic applications. Surface-based parameterization techniques, such as the B-spline surface \cite{b2,b3,b4,b5,b6} and other models \cite{b7}, can accurately fit complex geometries and curvature in point clouds. They typically rely on region-growing \cite{b5, b8} or supervoxel segmentation \cite{b9} for curved surface detection, which needs careful parameter tuning. These approaches require multiple passes over the data, making real-time operations on low-power onboard computers difficult. In addition to curvature fitting, B-spline surfaces demand boundary description, involving either manual selection of control points \cite{b5} or computationally expensive automatic refinement \cite{b6}.

\begin{figure}[t]
      \centering
      \includegraphics[scale=0.45]{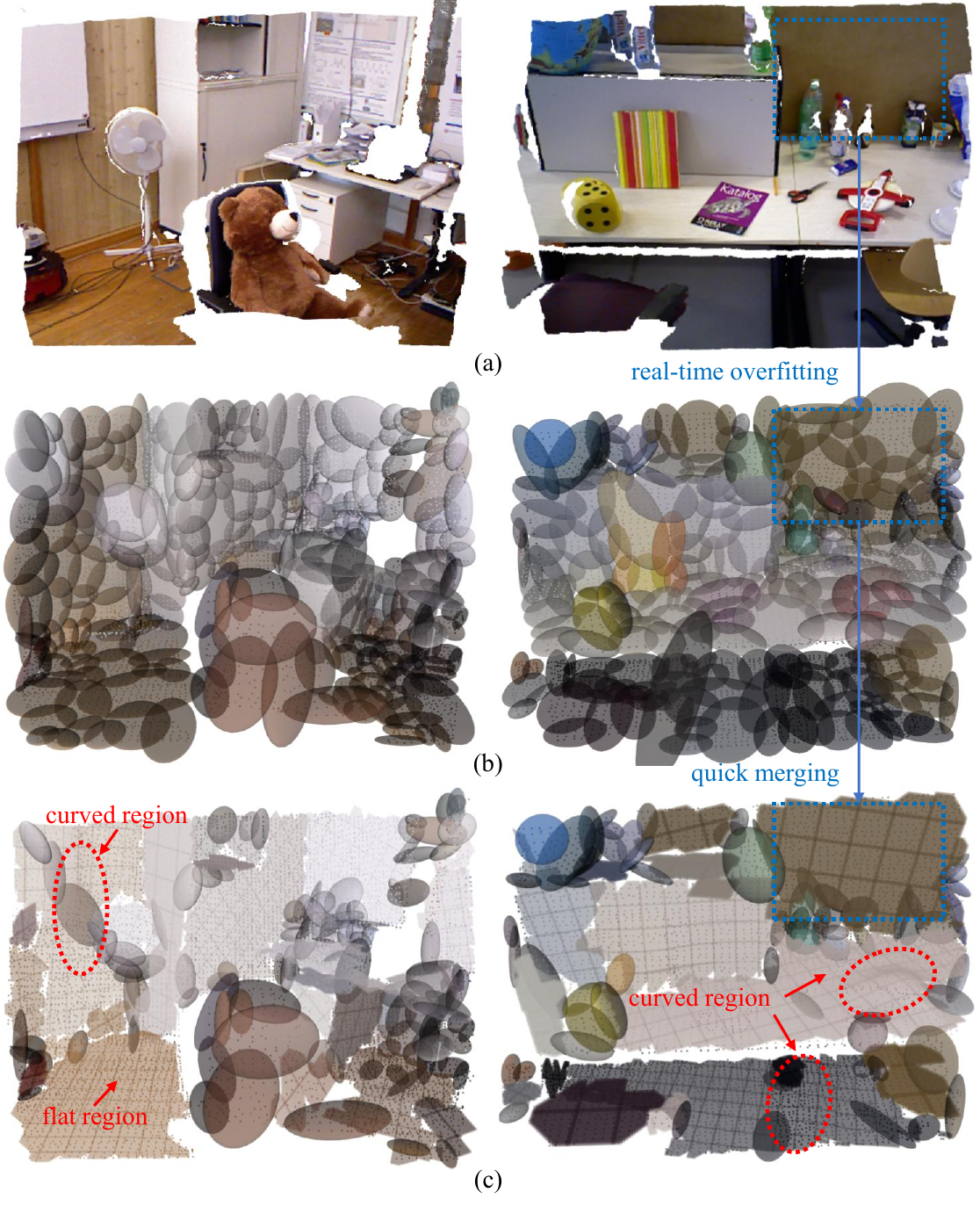}
      \caption{Demonstration of multi-model representation. (a), (b), and (c) show the original frames, GMM-based segmentations and multi-model representations, respectively. Flat regions are represented by planes, curved regions by B-spline surfaces, and unstructured regions by Gaussian distributions.}
      \label{figure_1_first_one}
\end{figure}

\begin{figure}[t!]
      \centering
      \vspace{0.3cm}
      \includegraphics[scale=0.43]{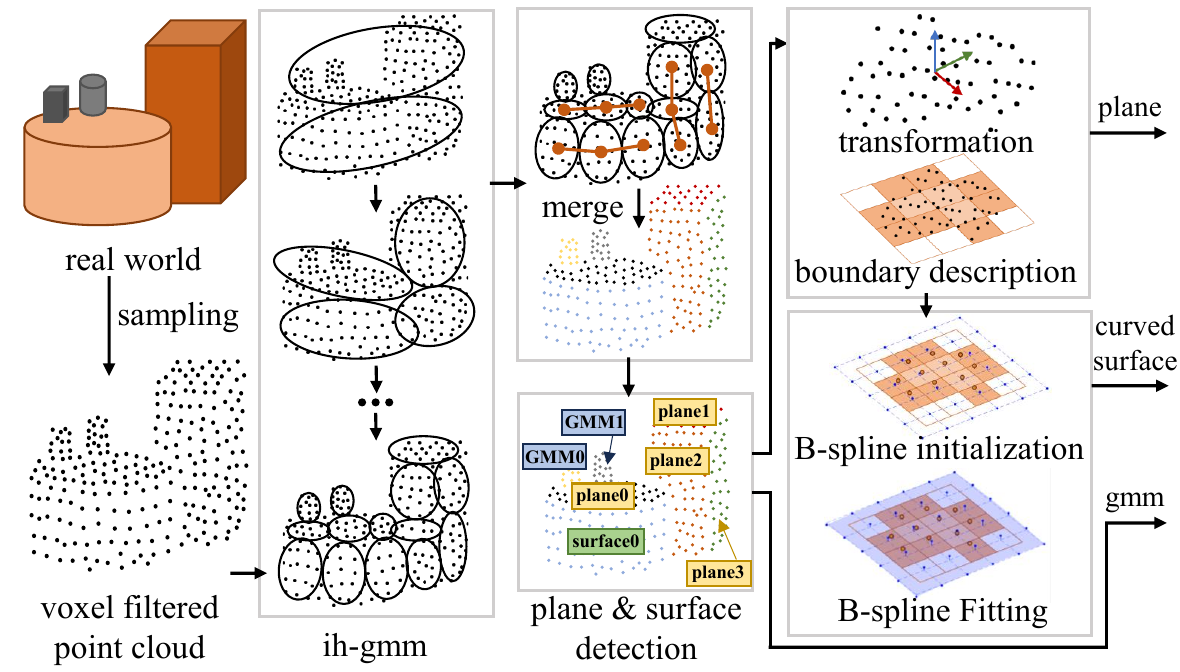}
      \vspace{-0.3cm}
      \caption{The overall process of the proposed method. A point cloud sampled from the real world is first voxel-filtered, and then over-segmented using an integrated hierarchical clustering method. Flat clusters are extracted and merged together and three types of clusters are obtained: GMMs, planes, and surfaces. Points belonging to planes and surfaces are transformed into their respective local coordinate systems, and their boundaries are described. For curved surfaces, B-spline surface fitting is applied to model curvature.}
      \label{figure_method}
\end{figure}
% 把点的大小修改大一些

\begin{algorithm}[t]
	\caption{Multi-model Parametric Representation}
	{\small
	\mbox{\textbf{Input:} Point Matrix $\mathbf{P}$} \\
	\mbox{\textbf{Output:} Model Parameters $\boldsymbol{\Theta}^\mathcal{G}$, $\boldsymbol{\Theta}^\mathcal{P}$, $\boldsymbol{\Theta}^\mathcal{S}$}}
        \vspace{-0.4cm}
	\begin{algorithmic}[1]
	{
      \small
	\STATE {$\mathcal{M} \triangleq \{\mathbf{P}_0,\mathbf{P}_1,\mathbf{P}_2\dots \}$ }
	\STATE $\textbf{procedure}$ H{\scriptsize IERARCHICAL}\_M{\scriptsize ULTIMODEL}\_R{\scriptsize EPRESENTATION}$(\mathbf{P})$
      \STATE \hspace{0.8em} $\mathcal{M} \leftarrow \text{IntegratedHierarchicalClustering}(\mathbf{P})$ \ref{sub::ihgmm}
      \STATE \hspace{0.8em} $\{\mathcal{M}^\mathcal{G},\mathcal{M}^\mathcal{P},\mathcal{M}^\mathcal{S}\} \leftarrow \text{MergeAndDetection}(\mathcal{M})$ \ref{sub::merge}
      \STATE \hspace{0.8em} {$\textbf{for } i\leftarrow0;i<\mathcal{M}^\mathcal{G}.\text{size}()-1;i\leftarrow i+1$}
      \STATE \hspace{1.6em} $\boldsymbol{\Theta}^\mathcal{G}.\text{pushback}(\text{GaussianEstimate}(\mathbf{P}^\mathcal{G}_i))$
      \STATE \hspace{0.8em} {$\textbf{end for} $}
      \STATE \hspace{0.8em} {$\textbf{for } j\leftarrow0;j<\mathcal{M}^\mathcal{P}.\text{size}()-1;j\leftarrow j+1$}
      \STATE \hspace{1.6em} $\{\theta^\mathcal{P}.mean,\theta^\mathcal{P}.cov,\tilde{\mathbf{P}}^\mathcal{P}_j\}\leftarrow\text{Transform}(\mathbf{P}^\mathcal{P}_j)$ \ref{sub::plane}
      \STATE \hspace{1.6em} $\theta^\mathcal{P}.bdy\leftarrow\text{BoundaryCal}(\tilde{\mathbf{P}}^\mathcal{P}_j)$ \ref{sub::boundary}
      \STATE \hspace{1.6em} $\boldsymbol{\Theta}^\mathcal{P}.\text{pushback}(\theta^\mathcal{P})$
      \STATE \hspace{0.8em} {$\textbf{end for} $}
      \STATE \hspace{0.8em} {$\textbf{for } k\leftarrow0;k<\mathcal{M}^\mathcal{S}.\text{size}()-1;k\leftarrow k+1$}
      \STATE \hspace{1.6em} $\{\theta^\mathcal{S}.mean,\theta^\mathcal{S}.cov,\tilde{\mathbf{P}}^\mathcal{S}_k\}\leftarrow\text{Transform}(\mathbf{P}^\mathcal{S}_k)$ \ref{sub::plane}
      \STATE \hspace{1.6em} $\theta^\mathcal{S}.bdy\leftarrow\text{BoundaryCal}(\tilde{\mathbf{P}}^\mathcal{S}_k)$ \ref{sub::boundary}
      \STATE \hspace{1.6em} $\theta^\mathcal{S}.control\_points \leftarrow \text{BsplineFitting}(\tilde{\mathbf{P}}^\mathcal{S}_k,\theta^\mathcal{S}.bdy)$ \ref{section::bspline}
      \STATE \hspace{1.6em} $\boldsymbol{\Theta}^\mathcal{S}.\text{pushback}(\theta^\mathcal{S})$
      
      \STATE \hspace{0.8em} {$\textbf{end for} $}
      \STATE \hspace{0.8em} {$\textbf{return } \boldsymbol{\Theta}^\mathcal{G}\text{,} \boldsymbol{\Theta}^\mathcal{P}\text{,}\boldsymbol{\Theta}^\mathcal{S}$}
	\STATE $\textbf{end procedure}$
      % \STATE
	}
	\end{algorithmic}
	\label{algo_w}
\end{algorithm}

In contrast to computationally expensive high-precision models, some real-time parametric approaches provide accuracy that is sufficient for robotic tasks.
Voxel maps and octree maps \cite{c0,c1} are widely used for obstacle avoidance, representing point clouds as collections of cubes. However, they are essentially a form of point cloud down-sampling, making it difficult to extract geometric features from the environment.
Parametric representations based on planes and triangular meshes \cite{c2,c3,c4} are well-suited for modeling planar regions in the environment but fail to approximate curved surfaces with a small number of geometric primitives.
Quadratic surface-based methods \cite{a1,c5,c6} can fit both planar and curved surfaces. Unfortunately, due to their implicit formulation and the presence of multiple solutions, representing point clouds with these methods makes boundary description cumbersome. Except for voxel maps, plane- and surface-based methods can only fit structured regions in the point cloud. For unstructured point clouds, balancing accuracy and memory consumption is challenging.

To mitigate this issue, highly real-time Gaussian mixture models (GMMs) have been widely proposed in recent years \cite{d0,d1,d2,d3,d4}. The construction of GMM maps can be classified into two main approaches. The first is optimization-based, typically relying on the expectation-maximization (EM) algorithm \cite{d0,d1}. The second involves region-partitioning and merging \cite{d2,d3,d4}. Several studies \cite{a3,d5} have extended these local mapping approaches to achieve global map construction. Such methods enable real-time model representations on robots with limited computational resources. However, ellipsoids struggle to accurately fit straight boundaries and sharp corners. Therefore, without increasing the number of distributions, such methods suffer from a limited accuracy ceiling.

In summary, GMM-based approaches fit unstructured point clouds well but exhibit lower accuracy for structured regions. Plane-based representations achieve higher accuracy for flat regions but are not memory-efficient for curved regions. Curvature can be better fitted by curved surfaces.
Combining the strengths of planes, curved surfaces and GMMs enables a multi-model point cloud representation: GMMs handle unstructured regions, while planes or curved surfaces model highly structured areas. Existing multi-model methods are mainly applied in SLAM systems. \cite{a1} represents point clouds using quadratic surfaces and degenerate quadrics, i.e., Gaussian distributions. This approach replaces surface detection with point cloud segmentation \cite{e0}, achieving real-time performance at the cost of low detection rates. Other multi-model representations, based on lines, planes and cylinders, do not effectively fit coarse regions \cite{a0,e1,e2,e3}.
The main challenge is how to represent curved surfaces in a multi-model representation and to develop a real-time and suitable surface detection for such surface models.

To tackle the problem, this paper proposes a multi-model parametric representation of point clouds with real-time surface detection and fitting (Fig. \ref{figure_1_first_one}). The point cloud is first represented using GMMs. Next, flat point cloud clusters are extracted from GMMs and merged into larger surfaces according to their normal vectors and distances between them. The merged point clouds can be fitted with planes and B-spline surfaces. Finally, the planes and surfaces are projected onto a two-dimensional mesh, with their boundaries determined by constraining each voxel's extents. Our main contributions are as follows:

\begin{enumerate}
\item Our approach improves the \textbf{robustness} and \textbf{efficiency} of curved surface detection. Compared with previous methods for B-spline surface detection, our approach achieves an efficiency improvement of \textbf{3.78} times.

\item This work presents a multi-model parametric representation framework that effectively integrates Gaussian distributions, planes, and B-spline surfaces, choosing the most appropriate model based on each region's geometric characteristics. The proposed method achieves a \textbf{twofold} improvement in accuracy over Gaussian mixture models with negligible efficiency loss.
\end{enumerate}

\section {Algorithm Overview}

The system pipeline is shown in Fig. \ref{figure_method} and summarized in Algorithm \ref{algo_w}. The algorithm takes point cloud coordinates as input and outputs the parameters of each model. Lines 3-4 cover integrated hierarchical clustering for GMM (Section \ref{sub::ihgmm}) and the subsequent merging and detection (Section \ref{sub::merge}). In Lines 9-10, each detected plane is transformed into its local coordinate system and its boundary is determined (Sections \ref{sub::plane} and \ref{sub::boundary}). Curved surfaces are fitted with B-spline surfaces in Line 16 (Section \ref{section::bspline}).

\section {GMM-Based Plane and Surface Detection}
\label{section::ihgmm}
In this section, a method for plane and surface detection is described. In Section \ref{sub::ihgmm}, we present an integrated hierarchical clustering approach based on GMM for segmenting the point cloud. Then, the merging strategy of clusters for plane and surface detection is introduced in Section \ref{sub::merge}.

\subsection {Integrated Hierarchical Clustering Based on GMM}
\label{sub::ihgmm}
An integrated hierarchical approach is used for fitting the point clouds with planar geometric primitives. In the previous work \cite{d0}, the point cloud is segmented hierarchically by K-means \cite{f0} and EH-GMM \cite{f1} algorithms to boost efficiency and the segmentation is terminated by an evaluation index. The evaluation index considers maximum likelihood function, model complexity and reconstruction accuracy. However, it's not suitable for this work because the hierarchical approach may terminate before the point cloud is segmented into clusters shaped like planar geometric primitives. Therefore, a new evaluation strategy for termination is proposed.

The point cloud is filtered with $a_\text{voxel}$-size to decrease computational load. A point matrix representing the voxel-filtered points is defined as $\mathbf{P}=\left[\mathbf{p}_{0},\ldots,\mathbf{p}_{n}, \ldots,\mathbf{p}_{N-1}\right]^T$, where $\mathbf{p}_{n} \in \mathbb{R}^{3 \times 1}$ is the 3D coordinate vector of $n$th point. A one-to-two partitioning operation is denoted as
\begin{equation}
\begin{alignedat}{2}
&\mathcal{C}: \bigcup_{N=4}^{\infty} \mathbb{R}^{N \times 3} \to \bigcup_{N'=2}^{\infty} \mathbb{R}^{N' \times 3} \times \bigcup_{N''=2}^{\infty} \mathbb{R}^{N'' \times 3} \\
&\mathcal{C}\left(\mathbf{P}\right) = \left(\mathbf{P}',\mathbf{P}''\right), N=N'+N''
\label{equ 1}
\end{alignedat}
\end{equation}
where each $\mathbf{p}_n^T$ is a row of $\mathbf{P}'$ or $\mathbf{P}''$. The assignment of row $\mathbf{p}_n^T$ is determined by a clustering algorithm selected by % where $\mathbf{c} = \left[c_{0},\ldots,c_{n}, \ldots,c_{N-1}\right]^T$ represents the index of the set to which each point belongs. New generated two point matrices are $\mathbf{P}'\in\mathbb{R}^{N' \times 3}$ and $\mathbf{P}''\in\mathbb{R}^{(N-N') \times 3}$. $\forall \mathbf{p}'_{n'}{}^T$ is a line of $\mathbf{P}'$, $\mathbf{p}'_{n'}=\mathbf{p}_{n}$ and $c_n=0$. $\forall \mathbf{p}''_{m'}{}^T$ in $\mathbf{P}''$, $\mathbf{p}''_{m'}=\mathbf{p}_{m}$, $c_m=1$. 
\begin{equation}
\mathcal{C}\left(\mathbf{P}\right)=\left\{\begin{array}{ll}
\mathcal{C}_{\text{K-means}}  \left(\mathbf{P}\right),& N>N_\text{em} \\
\mathcal{C}_{\text{EM-GMM}}  \left(\mathbf{P}\right),& N \leq N_\text{em}
\end{array}\right.
\label{equ 2}
\end{equation}
where $N_\text{em}$ is a parameter set manually. It limits the number of points in clusters segmented by the EM-GMM algorithm to increase efficiency. To represent a segmentation of the point cloud, the $I$-th set in iteration is defined as
\begin{equation}
{}^I\mathcal{M} = \left\{ {}^1\mathbf{P},{}^2\mathbf{P},\ldots,{}^I\mathbf{P}\right\},{}^i\mathbf{P} \in \mathbb{R}^{N_i \times 3}, \; i = 1,\ldots,I 
\end{equation}
where ${}^i\mathbf{P}$ is the point matrix of $i$-th cluster in ${}^I\mathcal{M}$. The set is initialized to ${}^1\mathcal{M} = \left\{ \mathbf{P}\right\}$, and the iteration proceeds as:
\begin{equation}
\begin{aligned}
&\text{if}\ f\left( {}^i\mathbf{P}\right) = 0, \mathcal{C}\left({}^i\mathbf{P}\right) = ({}^i\mathbf{P}',{}^i\mathbf{P}'')\\
&{}^{I+1}\mathcal{M} = \left({}^{I}\mathcal{M} \text{\textbackslash} {}^i\mathbf{P} \right) \cup \left\{ {}^i\mathbf{P}', {}^i\mathbf{P}'' \right\} \\
% x &= y \cdot z
\end{aligned}
\label{iteration}
\end{equation}
where $f:\bigcup_{N=2}^{\infty}\mathbb{R}^{N\times 3}\rightarrow\{0,1\}$ represents whether the point cluster ${}^i\mathbf{P}$ needs to be further segmented.

\begin{figure}[t]
      \centering
      \vspace{0.2cm}
      \includegraphics[scale=0.52]{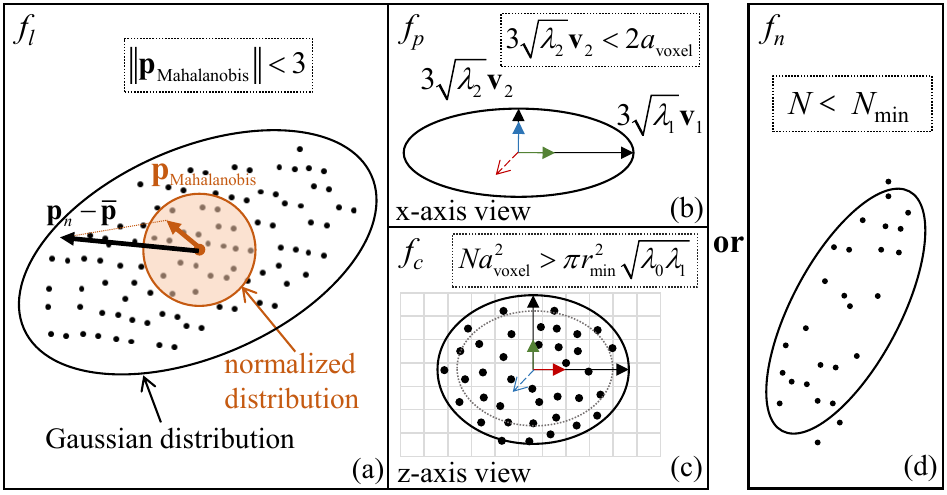}
      \vspace{-0.2cm}
      \caption{Termination judgment for integrated hierarchical clustering based on GMM. The Mahalanobis distance is the normalized distance in a Gaussian distribution. If a point lies outside the three-sigma bound, the distribution is considered non-compact and requires further partitioning. In the x-axis view, the principal axis of ellipse along z-axis should be sufficiently short to ensure that the cluster can be fitted as a planar primitive. If this condition is satisfied, the projection of the cluster onto the xy-plane should be dense. If neither of the above conditions is met, only a constraint is imposed to prevent overfitting.}
      \label{figure_ihgmm}
\end{figure}

The definition of $f$ is described in detail as follows. By calculating the mean vector ${}^{i}\bar{\mathbf{p}} = {N_i}^{-1} \sum_{n=0}^{N_i-1} {}^{i}\mathbf{p}_n$ and covariance matrix ${}^i\mathbf{\Sigma} = {(N_i-1)}^{-1} \sum_{n=0}^{N_i-1} ({}^{i}\mathbf{p}_n - {}^{i}\bar{\mathbf{p}})({}^{i}\mathbf{p}_n - {}^{i}\bar{\mathbf{p}})^T$, the eigenvalues ${}^i\lambda_0, {}^i\lambda_1, {}^i\lambda_2$ and normalized eigenvectors ${}^i\mathbf{v}_0, {}^i\mathbf{v}_1, {}^i\mathbf{v}_2$ of covariance matrix can be obtained by ${}^i\mathbf{\Sigma} {}^i\mathbf{v}_j = {}^i\lambda_j {}^i\mathbf{v}_j, \ j = 0,1,2$, where the eigenvalues satisfy $\lambda_0 \ge \lambda_1 \ge \lambda_2$. According to the classic principal component analysis (PCA) \cite{f2}, the three-sigma boundary of the points forms an ellipsoid with principal axes ${}^i\mathbf{x} = 3\sqrt{{}^i\lambda_0}{}^i\mathbf{v}_0, {}^i\mathbf{y} = 3\sqrt{{}^i\lambda_1}{}^i\mathbf{v}_1, {}^i\mathbf{z} = 3\sqrt{{}^i\lambda_2}{}^i\mathbf{v}_2$. Based on the ellipsoid, $f$ is defined as $f({}^i\mathbf{P}) = \min\{1,f_l({}^i\mathbf{P}) \cdot f_p({}^i\mathbf{P}) \cdot f_c({}^i\mathbf{P}) + f_n({}^i\mathbf{P})\}$, with each item illustrated in Fig. \ref{figure_ihgmm}. To ensure complete envelopes of Gaussian distributions, the Mahalanobis distance judgment $f_l$ is denoted as
\begin{equation}
\begin{aligned}
&f_l({}^i\mathbf{P}) = \prod _{n=0}^{N_i-1}s_l({}^i\mathbf{p}_n) \\
&s_l({}^i\mathbf{p}_n) = 
      \begin{cases}
      1, & ({}^i\mathbf{p}_n-{}^{i}\bar{\mathbf{p}})^T[{}^i\mathbf{\Sigma}]^{-1}({}^i\mathbf{p}_n-{}^{i}\bar{\mathbf{p}}) < 3^2 \\
      0, & \text{else} \\
      \end{cases}
\end{aligned}
\end{equation}
where $({}^i\mathbf{p}_n-{}^{i}\bar{\mathbf{p}})^T[{}^i\mathbf{\Sigma}]^{-1}({}^i\mathbf{p}_n-{}^{i}\bar{\mathbf{p}})$ is the square of the Mahalanobis distance from ${}^i\mathbf{p}_n$ to its mean vector ${}^{i}\bar{\mathbf{p}}$. The judgment of flatness $f_p$ is defined as
\begin{equation}
f_p({}^i\mathbf{P}) = 1 \text{ if } 6\sqrt{{}^i\lambda_2}<a_\text{voxel}, \,\text{else } 0
\end{equation}

The density limitation $f_c$ is described as
\begin{equation}
f_c({}^i\mathbf{P}) = 1 \text{ if } N_ia_\text{voxel}^2>\pi r^2_\text{min}\sqrt{{}^i\lambda_0{}^i\lambda_1}, \,\text{else } 0
\end{equation}
where $r_\text{min}\in[2,3]$ and is set manually. To prevent overfitting, the number of points in a cluster is constrained by
\begin{equation}
f_n({}^i\mathbf{P}) = 1 \text{ if } N_i < N_\text{min}, \,\text{else } 0
\end{equation}
where $N_\text{min}$ is the minimum allowed number of points in a cluster. The integrated hierarchical clustering is completed by repeating (\ref{iteration}) until $\forall\ {}^i\mathbf{P} \in {}^I\mathcal{M},f\left( {}^i\mathbf{P}\right) = 1$.

\subsection {Merging of Clusters for Plane and Surface Detection}
\label{sub::merge}
After the GMM clustering is completed, the flat point clusters can be extracted for further merging. Formally:
\begin{equation}
      \begin{aligned}
      &{}^I\mathcal{M} = \mathcal{M}^\mathcal{G} \cup \mathcal{M}^\mathcal{B},\ \mathcal{M}^\mathcal{G} \cap \mathcal{M}^\mathcal{B}=\varnothing \\
      &\forall\ {}^i\mathbf{P}^\mathcal{G} \in \mathcal{M}^\mathcal{G}, f_l({}^i\mathbf{P}^\mathcal{G}) \cdot f_p({}^i\mathbf{P}^\mathcal{G}) \cdot f_c({}^i\mathbf{P}^\mathcal{G})=0 \\
      &\forall\ {}^j\mathbf{P}^\mathcal{B} \in \mathcal{M}^\mathcal{B}, f_l({}^j\mathbf{P}^\mathcal{B}) \cdot f_p({}^j\mathbf{P}^\mathcal{B}) \cdot f_c({}^j\mathbf{P}^\mathcal{B})=1
      \end{aligned}
\end{equation}
where $\mathcal{M}^\mathcal{G}$ represents the set of clusters that should be fitted by Gaussian distributions and clusters in $\mathcal{M}^\mathcal{B}$ can be fitted by planar geometric primitives.

To extract planes and surfaces, planar-shaped clusters are merged by concatenating their corresponding point matrices. The judgment function is denoted as
\begin{equation}
d: \bigcup_{N_i=2}^{\infty} \mathbb{R}^{N_i \times 3} \times \bigcup_{N_j=2}^{\infty} \mathbb{R}^{N_j \times 3} \to \left\{0,1\right\}\\
\end{equation}

\begin{figure}[b]
      \centering
      \includegraphics[scale=0.50]{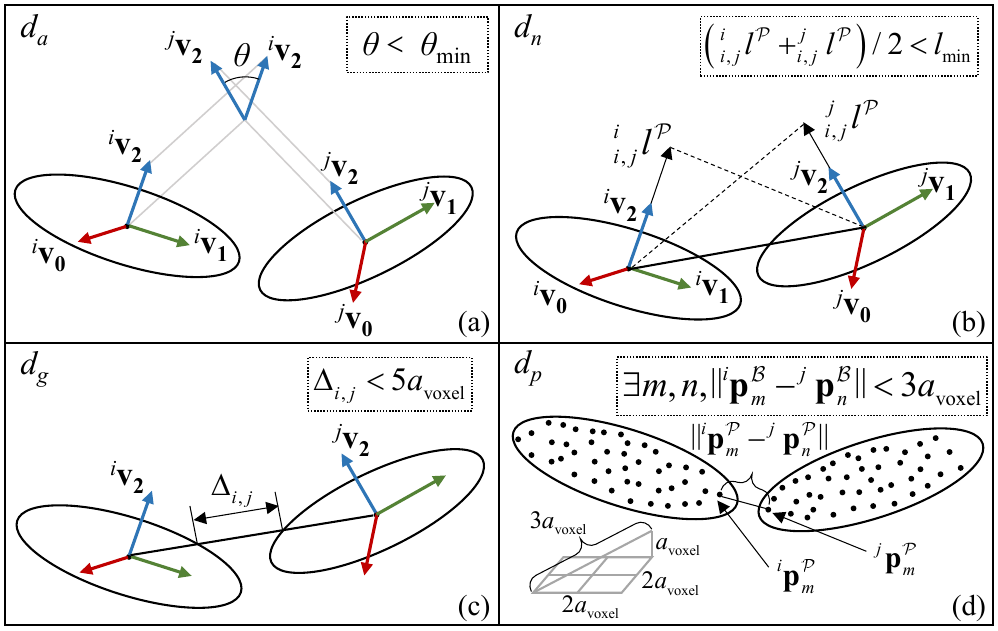}
      \caption{Merging judgment between two planar geometric primitives described by Gaussian distributions. Firstly, the two planes are required to be parallel.
      The normal-direction distance between the centroids of two parallel planes should be less than the size of a voxel. Considering possible fitting errors, a distance of less than twice the voxel size may also be accepted. Once these conditions are satisfied, the two planes should be adjacent. A simple geometric distance is computed using the Gaussian distributions to filter out pairs that are far apart. Finally, the point-to-point distances between the two clusters are computed to determine adjacency.}
      \label{figure_merge}
      % first line use voxel filtering
      % second line use seg-point with differernt colors 
      % third line use now figure (label the lines gmms and bsplines)
\end{figure}

If $d({}^i\mathbf{P}^\mathcal{B},{}^j\mathbf{P}^\mathcal{B}) = 1$, the two clusters can be merged into a larger one. $d$ is the product of multiple symbolic functions, i.e., $d = d_a \cdot d_n \cdot d_g \cdot d_p$, and its brief description is shown in Fig. \ref{figure_merge}. $d_a$ requires that the normal directions of the two clusters are similar, which is expressed as
\begin{equation}
d_a({}^i\mathbf{P}^\mathcal{B},{}^j\mathbf{P}^\mathcal{B})=1 \text{ if } \left| {}^i\mathbf{v}_2^\mathcal{B} \cdot {}^j\mathbf{v}_2^\mathcal{B} \right| > \cos (\theta_\text{min}), \,\text{else }0
\end{equation}
where $\theta_\text{min}$ is the smallest angle can be tolerated. The vector from the mean of ${}^i\mathbf{P}^\mathcal{B}$ to mean of ${}^j\mathbf{P}^\mathcal{B}$ is $\Delta {}_{i,j}\bar{\mathbf{p}}^\mathcal{B}= {}^{i}\bar{\mathbf{p}}^\mathcal{B}-{}^{j}\bar{\mathbf{p}}^\mathcal{B}$, projecting it onto ${}^i\mathbf{v}_2^\mathcal{B}$ can obtain the distance between the two means in the normal direction of ${}^i\mathbf{P}^\mathcal{B}$, i.e., ${}_{i,j}^il^\mathcal{B}=\|{}^i\mathbf{v}_2^\mathcal{B} \cdot \Delta {}^{i}_{j}\bar{\mathbf{p}}^\mathcal{B}\|$. According to this, the distance in normal direction judgment $d_n$ is described as
\begin{equation}
d_n({}^i\mathbf{P}^\mathcal{B},{}^j\mathbf{P}^\mathcal{B})=1 \text{ if } ({}_{i,j}^il^\mathcal{B}+{}_{i,j}^jl^\mathcal{B})/2<l_\text{min}, \,\text{else }0
\end{equation}
where $l_\text{min}$ is set in $[a_\text{voxel},2a_\text{voxel}]$. $d_g$ is denoted as the judgment of Euclidean distance between two clusters. The distance is defined as the center-to-center distance, excluding the segments lying within the three-sigma boundaries of the two distributions. The normalized vector of the connecting line is $\Delta {}_{i,j}\bar{\mathbf{p}}^\mathcal{B}/\|\Delta {}_{i,j}\bar{\mathbf{p}}^\mathcal{B} \|$, rotating it into the space spanned by the three-axis orthogonal basis of $i$-th distribution, i.e., $\Delta _{i,j}\bar{\mathbf{p}}^\mathcal{B}_\mathbf{v}=  [{}^i\mathbf{Q}^\mathcal{B}]^T\Delta _{i,j}\bar{\mathbf{p}}^\mathcal{B}/\|\Delta {}_{i,j}\bar{\mathbf{p}}^\mathcal{B} \|$, where ${}^i\mathbf{Q}^\mathcal{B} = [{}^i\mathbf{v}_0^\mathcal{B},{}^i\mathbf{v}_1^\mathcal{B},{}^i\mathbf{v}_2^\mathcal{B}]$ is an orthogonal matrix. Scaling $\Delta _{i,j}\bar{\mathbf{p}}^\mathcal{B}_\mathbf{v}$ to the boundary of three-sigma, the length within the boundary of $i$-th distribution is calculated by ${}_{i,j}^il^\mathcal{B}_\mathbf{v} = 3[\sqrt{{}^i\lambda_0^\mathcal{B}}, \sqrt{{}^i\lambda_1^\mathcal{B}}, \sqrt{{}^i\lambda_2^\mathcal{B}}]\Delta _{i,j}\bar{\mathbf{p}}^\mathcal{B}_\mathbf{v}$. Therefore, the distance between the two boundaries is defined as $\Delta_{i,j} = \|\Delta {}_{i,j}\bar{\mathbf{p}}^\mathcal{B} \|-{}_{i,j}^il^\mathcal{B}_\mathbf{v}-{}_{i,j}^jl^\mathcal{B}_\mathbf{v}$, and $d_g$ can be described as
\begin{equation}
d_g({}^i\mathbf{P}^\mathcal{B},{}^j\mathbf{P}^\mathcal{B}) = 1 \text{ if } \Delta_{i,j} <5a_\text{voxel}, \,\text{else }0
\end{equation}

$d_g$ can be regarded as a preliminary filter for $d_p$. The computation of $d_p({}^i\mathbf{P}^\mathcal{B},{}^j\mathbf{P}^\mathcal{B})$ traverses the points in the two clusters to calculate the length, i.e.,
\begin{equation}
d_p = 1 \text{ if } \exists m,n, \|{}^{i}\mathbf{p}^\mathcal{B}_m - {}^{j}\mathbf{p}^\mathcal{B}_n\|<3a_\text{voxel} , \,\text{else }0
\end{equation}

After computing the merging judgment function between any two clusters, connected components can be identified if such relationships exist. The point matrices of the clusters within each connected component are concatenated by
\begin{equation}
\mathbf{P}^\text{new} = [\ldots;{}^i\mathbf{P}^\mathcal{B};{}^j\mathbf{P}^\mathcal{B};\ldots], \forall i, \exists j,d({}^i\mathbf{P}^\mathcal{B},{}^j\mathbf{P}^\mathcal{B}) = 1 
\end{equation}
where $\mathbf{P}^\text{new}$ is an example of concatenated matrices. A cluster that cannot be merged with others forms a separate component, i.e., its concatenated matrix is the point matrix itself. By the procedures above, a set of concatenated point matrices, called $\mathcal{M}^\text{new}$, is calculated. $\mathcal{M}^\text{new}$ is partitioned into merged plane set $\mathcal{M}^\mathcal{P}$ and curved surface set $\mathcal{M}^\mathcal{S}$, and elements in them satisfy
\begin{equation}
      \begin{aligned}
      & \mathcal{M}^\mathcal{P} \cup \mathcal{M}^\mathcal{P} = \mathcal{M}^\text{new}, \mathcal{M}^\mathcal{P} \cap \mathcal{M}^\mathcal{P} = \varnothing\\
      &\forall {}^k\mathbf{P}^\mathcal{P} \in \mathcal{M}^\mathcal{P} , f_p({}^k\mathbf{P}^\mathcal{P})=1\\
      &\forall {}^l\mathbf{P}^\mathcal{S} \in \mathcal{M}^\mathcal{S}, f_p({}^l\mathbf{P}^\mathcal{S})=0
      \end{aligned}
\end{equation}

The point cloud is segmented into a set of Gaussian distributions $\mathcal{M}^\mathcal{G}$, a plane set $\mathcal{M}^\mathcal{P}$, and a surface set $\mathcal{M}^\mathcal{S}$. For parameterization, each ${}^i\mathbf{P}^\mathcal{G} \in \mathcal{M}^\mathcal{G}$ can be represented by its mean and covariance matrix.

\section{Plane Fitting and Boundary Description}
\label{section::fit}
In this section, a method for plane fitting is described. In Section \ref{sub::plane}, we present a coordinate transformation for plane representation. Then, a method for boundary description based on 2D voxels is proposed in Section \ref{sub::boundary}.

\subsection{Coordinate Transformation and Plane Representation}
\label{sub::plane}
In this subsection, only points in ${}^k\mathbf{P}^\mathcal{P} \in \mathcal{M}^\mathcal{P}$ are considered. For clearer expressions, ${}^k\mathbf{P}^\mathcal{P}$ is simply written as $\mathbf{P}$.
For each $\mathbf{P}$, a mean vector $\bar{\mathbf{p}}$ and an orthogonal matrix consisting of three orthonormal bases $\mathbf{Q} = [\mathbf{v}_0,\mathbf{v}_1,\mathbf{v}_2]$ can be obtained. A coordinate transformation is performed on each point $\mathbf{p}_n \in \mathbf{P}$ by
\begin{equation}
      \tilde{\mathbf{p}}_n= [\mathbf{Q}]^T\left(\mathbf{p}_n-\bar{\mathbf{p}}\right), n =1,\ldots,N-1
\end{equation}
where $\tilde{\mathbf{p}}_n$ is a point in the space spanned by $\mathbf{Q}$. The plane's location and direction can be represented by $\bar{\mathbf{p}}$ and $\mathbf{Q}$.

\subsection{Voxel-Based Boundary Description}
\label{sub::boundary}

\begin{figure}[t!]
      \centering
      \vspace{0.2cm}
      \includegraphics[scale=0.62]{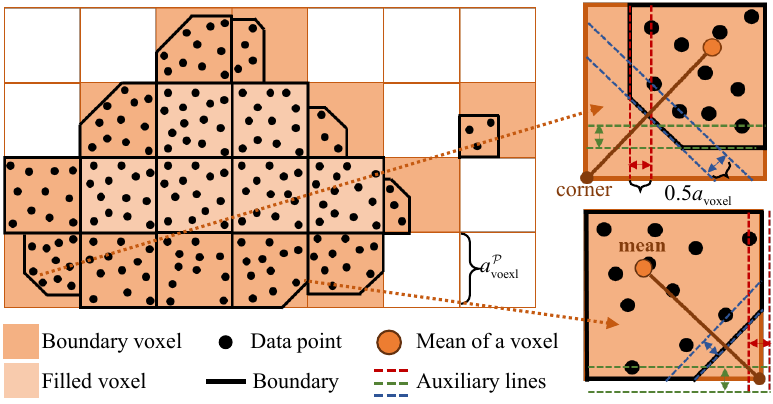}
      \vspace{-0.2cm}
      \caption{Boundary description is performed based on a 2D voxel map. The points are first partitioned into different voxels by overlaying a 2D mesh grid. The boundary voxels are then extracted for further boundary characterization. For each voxel, the mean of its points is computed to characterize the point distribution. Empty regions are removed applying line constraints across the grid in three directions: horizontal, $45^\circ$ diagonal, and vertical. These lines are offset from the data points by a margin of $a_\text{voxel}$ to avoid direct overlap.}
      \label{figure_boundary}
      % first line use voxel filtering
      % second line use seg-point with differernt colors 
      % third line use now figure (label the lines gmms and bsplines)
\end{figure}

An $a^\mathcal{P}_\text{voxel}$-sized 2D voxel map is calculated as illustrated in Fig. \ref{figure_boundary}. The points in each voxel are described by
\begin{equation}
\tilde{\mathbf{p}}_n \in \mathcal{V}_{i,j}, i = 0,\dots,N_x\!-\!1, j = 0,\dots,N_y\!-\!1
\end{equation}
where $\mathcal{V}_{i,j}$ is a set of points at location $(i,j)$. To record whether $\mathcal{V}_{i,j}=\varnothing$, $\mathbf{E}\in \{0,1\}^{N_x\times N_y}$ is defined. If $\mathcal{V}_{i,j}=\varnothing$, $\mathbf{E}_{i,j}=0$, else $\mathbf{E}_{i,j}=1$. To determine whether a voxel is at the boundary of a plane, $\mathbf{B}\in \{0,1\}^{N_x\times N_y}$ is defined. For a voxel at the edge of the voxel map (i.e., $i = 0$ or $i = N_x-1$ or $j = 0$ or $j = N_y-1$), if $\mathcal{V}_{i,j}\neq \varnothing$, $\mathbf{B}_{i,j}=1$. For a voxel not at the edge, if $\mathcal{V}_{i,j+1}=\varnothing\lor\mathcal{V}_{i+1,j}=\varnothing\lor\mathcal{V}_{i,j-1}=\varnothing\lor\mathcal{V}_{i-1,j}=\varnothing$ and $\mathcal{V}_{i,j}\neq \varnothing$, $\mathbf{B}_{i,j}=1$. Otherwise, $\mathbf{B}_{i,j}=0$. 

For each voxel at the boundary, three lines are used to remove empty regions as shown in Fig. \ref{figure_boundary}. The parameters of the lines in voxel $(i,j)$ are defined as $l^X_{i,j}$, $l^Y_{i,j}$ and $l^{XY}_{i,j}$. $l^X_{i,j}$ and $l^Y_{i,j}$ are optimized by
\begin{equation}
\begin{aligned}
&l^X_{i,j} = \mathop{\arg\min} \left|x(\tilde{\mathbf{p}}_n)-x'_{i,j}\right|-0.5a_\text{voxel}\\
&l^Y_{i,j} = \mathop{\arg\min} \left|y(\tilde{\mathbf{p}}_n)-y'_{i,j}\right|-0.5a_\text{voxel}\\
&s.t.\ \tilde{\mathbf{p}}_n \in \mathcal{V}_{i,j}
\end{aligned}
\end{equation}
where $(x'_{i,j},y'_{i,j})$ is the corner point without neighboring points in the voxel $(i,j)$, and $x(\mathbf{p})$ and $y(\mathbf{p})$ denote the respective coordinates of $\mathbf{p}$. $(x'_{i,j},y'_{i,j})$ is defined by
\begin{equation*}
\begin{aligned}
&(x'_{i,j},y'_{i,j})=\mathop{\arg\max}_{(x'_{i,j},y'_{i,j}) \in \mathcal{V}^\text{cor}_{i,j}} \|[x'_{i,j};y'_{i,j}] - \bar{\tilde{\mathbf{p}}}_{i,j}\| \\
&\mathcal{V}^{cor}_{i,j} = \{x^\text{min}_{i,j}, x^\text{max}_{i,j}\} \times \{y^\text{min}_{i,j}, y^\text{max}_{i,j}\} 
\end{aligned}
\end{equation*}
in which $x^\text{min}_{i,j}$, etc., represent the coordinates of corner points in voxel $(i,j)$ and $\bar{\tilde{\mathbf{p}}}_{i,j}$ is the mean. Similarly, $\forall \tilde{\mathbf{p}}_n \in \mathcal{V}_{i,j}$
\begin{equation}
l^{XY}_{i,j}\!=\!\mathop{\arg\min} \!\left|x(\tilde{\mathbf{p}}_n)\!-\!x'_{i,j}\right| \!+\! \left|y(\tilde{\mathbf{p}}_n)\!-\!y'_{i,j}\right|\!-\!\frac{\sqrt{2}}{2}a_\text{voxel}
\end{equation}

If the point $(x'_{i,j},y'_{i,j})$ is surrounded by non-empty voxels, the voxel is considered to contain no sparse regions, so $l^X_{i,j}$, $l^Y_{i,j}$ and $l^{XY}_{i,j}$ are all set to $0$. To parametrize a plane, several parameters are used, namely $\bar{\mathbf{p}}$, $\mathbf{Q}$, $a^\mathcal{P}_\text{voxel}$, $x^\text{min}_{0,0}$, $y^\text{min}_{0,0}$, $\mathbf{E}$ and $\mathbf{B}$ with corresponding $l^X_{i,j}$, $l^Y_{i,j}$ and $l^{XY}_{i,j}$.

\section{B-spline Surface Fitting}
\label{section::bspline}
In this section, we describe a strategy for B-spline surface fitting. The boundary description applies the method proposed in Section \ref{sub::boundary}. In Section \ref{sub::def_point}, the control points of B-spline surfaces are designed and initialized based on the voxels for boundary description. For the fitness of B-spline surface to data points, the control points of B-spline are optimized in Section \ref{sub::cal_point}.

\subsection{Initialization of the Control Points}
\label{sub::def_point}

\begin{figure}[b!]
      \centering
      \includegraphics[scale=0.52]{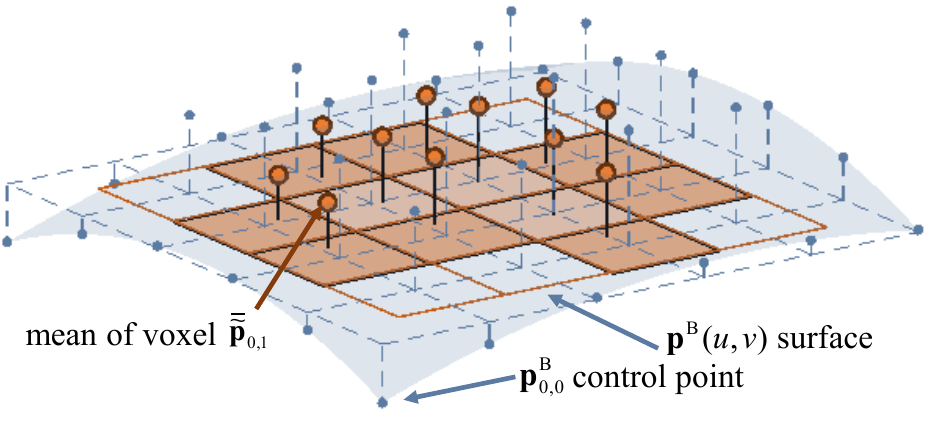}
      \caption{Computation of B-spline surface control points. Based on a two-dimensional grid, control points are initialized and these points can interpolate a surface. To ensure that the mean of points in each voxel lies on the surface, the coordinates of the control points are further optimized.}
      \label{figure_bspline}
      % first line use voxel filtering
      % second line use seg-point with differernt colors 
      % third line use now figure (label the lines gmms and bsplines)
\end{figure}

% Similar to Section \ref{sub::boundary}, sets $\mathcal{V}_{i,j}$ ($i = 0,\dots,N_x-1, j = 0,\dots,N_y-1$) are generated. 
Control points are applied by B-spline to adjust the curvature of surface. As illustrated in Fig. \ref{figure_bspline}, the $x$- and $y$-coordinates of the control points are located at the centers of the voxels in a 2D mesh (blue-gray), with an extra layer surrounding the boundary description mesh (brown). Control points are denoted by $\mathbf{p}^\text{B}_{\alpha,\beta}$ ($\alpha\!=\!0,\dots,N_x+1$, $\beta\!=\!0,\dots,N_y+1$). For the overlap region, i.e., $\alpha\!=\! 1,\dots,N_x$, $\beta\! = \!1,\dots,N_y$, $x$ and $y$ of control points are
\begin{equation}
      \begin{aligned}
      &x(\mathbf{p}^\text{B}_{\alpha,\beta}) = (x^\text{min}_{\alpha-1,\beta-1} + x^\text{max}_{\alpha-1,\beta-1})/2\\
      &y(\mathbf{p}^\text{B}_{\alpha,\beta}) = (y^\text{min}_{\alpha-1,\beta-1} + y^\text{max}_{\alpha-1,\beta-1})/2\\
      \end{aligned}
\end{equation}

The $z$-coordinate of $\mathbf{p}^\text{B}_{\alpha,\beta}$ is initialized by $z(\bar{\tilde{\mathbf{p}}}_{\alpha-1,\beta-1})$ and should be further optimized. For the extra layers, e.g., $\alpha\!=\! 0$, $\beta \!=\! 1,\dots,N_y$, the coordinates are
\begin{equation}
      \begin{aligned}
      &x(\mathbf{p}^\text{B}_{0,\beta}) = x^\text{min}_{0,\beta-1} - 0.5a^\mathcal{S}_\text{voxel},\ z(\mathbf{p}^\text{B}_{0,\beta}) = 0\\
      &y(\mathbf{p}^\text{B}_{0,\beta}) = (y^\text{min}_{0,\beta-1} + y^\text{max}_{0,\beta-1})/2
      \end{aligned}
      \label{control_init}
\end{equation}

For other extra layers, $\mathbf{p}^\text{B}_{\alpha,\beta}$ are initialized in a similar way as in (\ref{control_init}). Referring to \cite{b3}, the order of the B-spline curve is chosen as $T=3$ in both $x$ and $y$ directions. The knots vector in $x$ direction is defined as $\mathbf{u}^\text{x} \in [0,1]^{N_x+T+3}$ and $u^\text{x}_i$ is calculated by
\begin{equation}
      u^\text{x}_i=
      \begin{cases}
      0,                     &0\leq i\leq T\\
      \frac{i-T}{N_x+2-2T},  &T< i < N_x+2\\
      1,                     &N_x+2\leq i\leq N_x+T+2
      \end{cases}
\end{equation}

$\mathbf{u}^\text{x}$ in $y$ direction can be calculated in a same way.

\subsection{Optimization of Z-coordinates of Control Points}
\label{sub::cal_point}
The B-spline interpolation is essentially a weighted sum of the control points \cite{b3}, i.e.,
\begin{equation}
\mathbf{p}^\text{B}(u,v)=\sum_{\alpha=0}^{N_x+1}\sum_{\beta=0}^{N_y+1}W_{\alpha,\beta}(u,v)\mathbf{p}^\text{B}_{\alpha,\beta}, u,v \in [0,1]
\end{equation}
where $W_{\alpha,\beta}(u,v)$ is the weight of $\mathbf{p}^\text{B}_{\alpha,\beta}$ for calculating $\mathbf{p}^\text{B}(u,v)$. In this work, only $z$-axis has to be considered due to the definition of control points. As shown in Fig. \ref{figure_bspline}, by selecting appropriate pairs of $(u,v)$, the B-spline estimation of $z(\bar{\tilde{\mathbf{p}}}_{i,j})$, namely $z_{i,j}^\text{B}$, can be obtained by
\begin{equation}
      \begin{aligned}
      z_{i,j}^\text{B}=\!\sum_{\alpha=0}^{N_x+1}\sum_{\beta=0}^{N_y+1}W_{\alpha,\beta} \left( \frac{i+0.5}{N_x+1}+\frac{x(\bar{\tilde{\mathbf{p}}}_{i,j})-x^{\text{min}}_{i,j}}{a^\mathcal{S}_\text{voxel}},\right.\\
      \left.\frac{j+0.5}{N_y+1} + \frac{y(\bar{\tilde{\mathbf{p}}}_{i,j})-y^{\text{min}}_{i,j}}{a^\mathcal{S}_\text{voxel}} \right)z(\mathbf{p}_{\alpha,\beta})
      \end{aligned}
\end{equation}

For fitting, an unconstrained optimization problem with $z(\mathbf{p}_{\alpha,\beta})$ as optimization variables is formulated as
\begin{equation}
\min \sum_{i=0}^{N_x-1}\sum_{j=0}^{N_y-1}\|z(\bar{\tilde{\mathbf{p}}}_{i,j})-z_{i,j}^\text{B}\|^2
\end{equation}

Considering the linear relationships between $z_{i,j}^\text{B}$ and $z(\mathbf{p}^\text{B}_{\alpha,\beta})$, it is evident that this is a convex optimization problem. The cost function can achieve fast convergence through the gradient descent method. In summary, to represent a curved surface, only $z(\mathbf{p}^\text{B}_{\alpha,\beta})$ needs to be additionally recorded compared to the parameterization of a plane.

\section{Experiments}
\label{section::experiments}

\begin{figure}[t]
      \centering
      \vspace{0.2cm}
      \includegraphics[scale=0.50]{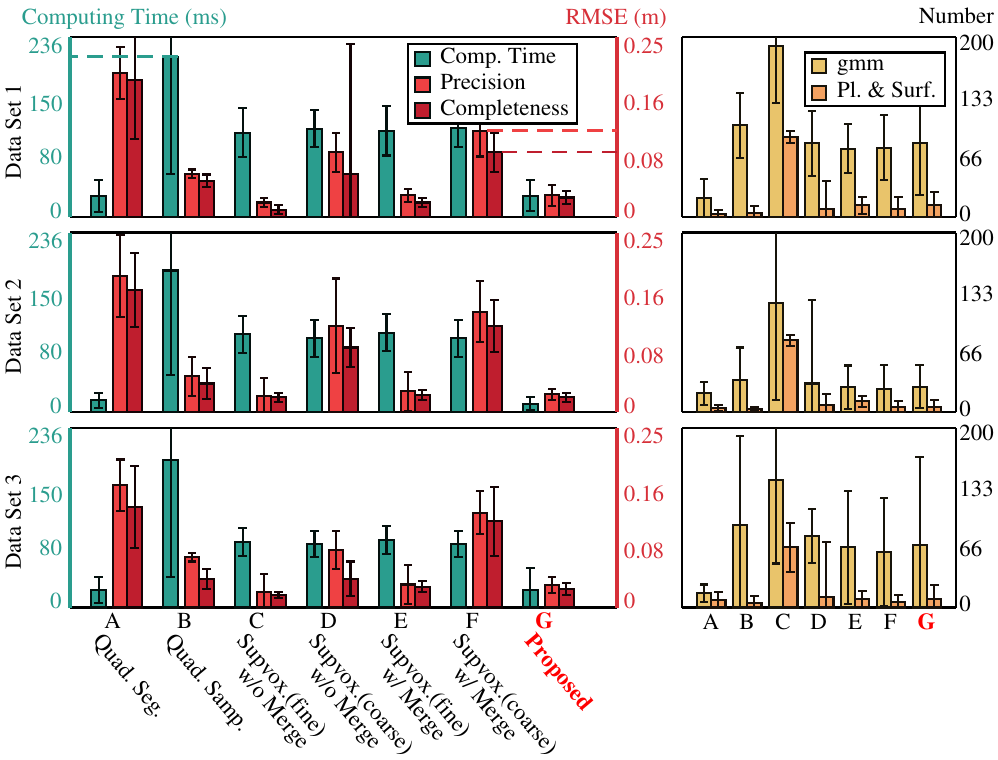}
      \vspace{-0.2cm}
      \caption{Results of the comparison between different detection methods. The green bars represent the computation time of each method, while the red bars indicate the RMSEs. In the right-hand figure, the yellow bars denote the number of fitted Gaussian distributions, and the orange bars represent the total number of detected planes and surfaces.}
      \label{figure_exp_surface_detec}
      % first line use voxel filtering
      % second line use seg-point with differernt colors 
      % third line use now figure (label the lines gmms and bsplines)
\end{figure}

\begin{figure}[t]
      \centering
      \vspace{0.2cm}
      \includegraphics[scale=0.45]{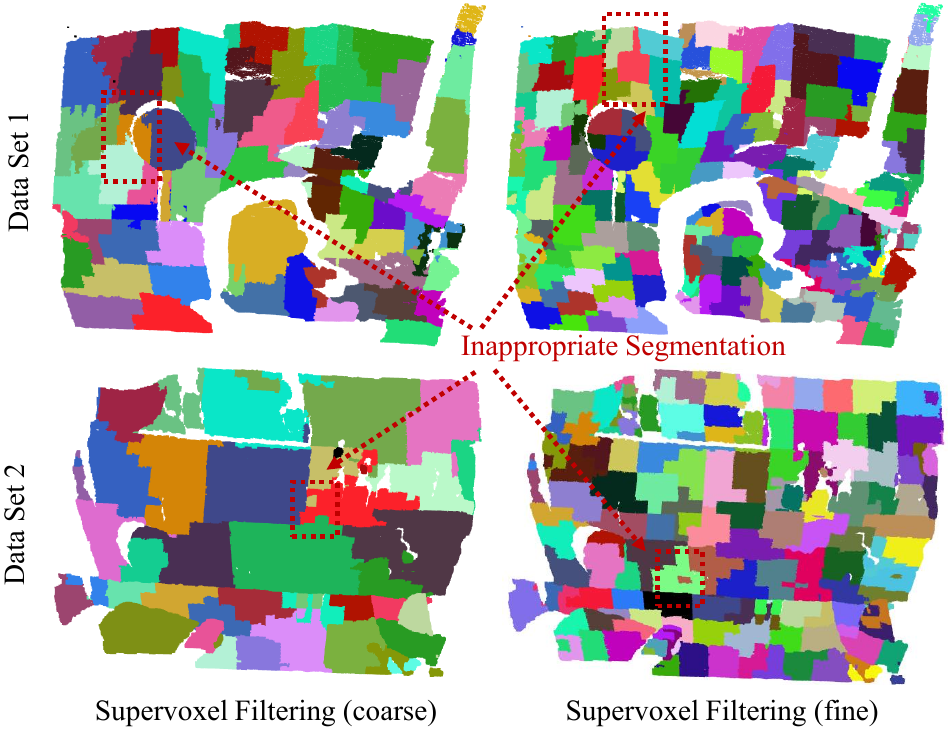}
      \vspace{-0.2cm}
      \caption{Demonstration of supervoxel filtering. The upper two figures highlight unreasonable segmentations caused by incorrectly placed supervoxels. The right one includes a corner. The lower two figures illustrate erroneous segmentations resulting from noise.}
      \label{figure_overfitting}
      % first line use voxel filtering
      % second line use seg-point with differernt colors 
      % third line use now figure (label the lines gmms and bsplines)
\end{figure}

In this section, we divide the experiments into two parts: detection of planes and surfaces, and model fitting. To demonstrate the superiority of our plane and surface detection method, we compare it with several existing approaches in Section \ref{sub::compare_surface}. To evaluate the trade-off between efficiency and accuracy, we compare our representation method against commonly used parametric models in Section \ref{sub::compare_model}. Finally, supplementary experiments on our approach are presented in Section \ref{sub::supplement}, and its characteristics are further analyzed.

For the proposed method, its parameters are set by a trial and error approach. $N_\text{em}$ is set as $200$, $r_\text{min}$ is $2$, $N_\text{min}$ is $40$ and $\theta_\text{min}$ is $15^{\circ}$. Three data sets are selected for testing, namely D1: TUM RGB-D freiburg1\_teddy, D2: TUM RGB-D freiburg3\_long\_office\_household\_validation \cite{data0}, and D3: ETH RGBD Voxblox \cite{data1}. In dataset D1, the size of the voxel filter $a_\text{voxel}\!=\!0.04$, and the other two are $0.03$. This depends on the scene size and density of the point clouds. The 2D voxel size for boundary description is set to $a^\mathcal{P}_\text{voxel}=5a_\text{voxel}$ for planes and $a^\mathcal{S}_\text{voxel}=3a_\text{voxel}$ for surfaces. All experiments are conducted on a commonly used low-power onboard computer, the Jetson Xavier NX (16GB).

A resampled point cloud can be generated from the parametric model representation. The precision of the generated point cloud is evaluated by the root mean squared error (RMSE) of the generated-to-real distance, i.e., the distance from each point in the generated point cloud to its nearest neighbor in the original point cloud. The completeness error is calculated in the reverse direction.

\subsection{Comparison of Surface Detection Methods}
\label{sub::compare_surface}

For the comparison of surface detection, we apply a point cloud segmentation method for quadratic fitting \cite{a1}, a sampling-based quadratic detection method \cite{c6} and typically used supervoxel filtering methods \cite{b2,b3,b4,b5,b6}. After the sampling method \cite{c6} is completed, to ensure that the remaining point clouds can also be fitted, the rest points are processed by the Gaussian clustering method of this work. In supervoxel filtering methods, the point cloud is first over-segmented and then the segmented clusters are merged. Since the approach is parameter-sensitive, the seed resolution of the supervoxel is set to either $0.15$ or $0.4$. These methods are only applied for detection, while the fitting is performed using the approach proposed in this work.

As shown in Fig. \ref{figure_exp_surface_detec}, the point cloud segmentation method suffers low accuracy because it focuses more on the Euclidean distance between point cloud clusters rather than the clusters' fitness to the surface. The sampling-based quadratic detection method can detect surfaces with high fitness from point clouds, but repetitive sampling operations are too time-consuming. Both methods have low detection capability.

\begin{table}[t!]
      \vspace{0.2cm}
  \caption{Comparison of Different Parametric Models}
  \centering
  \renewcommand{\arraystretch}{1}
  \begin{tabular}{ c p{0.28cm} p{0.28cm} p{0.28cm}    p{.28cm} p{.28cm} p{.28cm}    p{.28cm} p{.28cm} p{.28cm}}
    \toprule
      Evaluation & \multicolumn{3}{c}{Only Fitting} & \multicolumn{3}{c}{Precision} & \multicolumn{3}{c}{Completeness} \\ 
      Index & \multicolumn{3}{c}{Time (ms)} & \multicolumn{3}{c}{RMSE (m)} & \multicolumn{3}{c}{RMSE (m)} \\ 
      \midrule
      Models &D1&D2&D3&D1&D2&D3&D1&D2&D3 \\
      \midrule
      B-spline   & 621  & 428  & 591    & \underline{\bf{0.02}}  & \underline{\bf{0.01}} & \underline{\bf{0.02}}   & \underline{\bf{0.02}} & \underline{\bf{0.02}} & \underline{\bf{0.02}}\\ 
      GMM        & \ \ \underline{\bf{0}}&\ \ \underline{\bf{0}}&\ \ \bf{\underline{0}} & \underline{0.08}  & \underline{0.07} & \underline{0.07}   & \underline{0.06} & \underline{0.06} & \underline{0.07}\\ 
      Quadric+GMM   & \bf{1.02} & \bf{0.36} & \bf{0.71}   & \ \ --  & \ \ -- & \ \ --   & 0.19 & 0.17 & 0.14\\ 
\bf{Proposed}    & \underline{1.37} & \underline{0.41} & \underline{0.60}   & \bf{0.03}  & \bf{0.03} & \bf{0.03}   & \bf{0.03} & \bf{0.02} & \bf{0.03}\\ 
    \bottomrule
  \end{tabular}
  \label{tab:example}
\end{table}

\begin{figure}[t!]
      \centering
      \includegraphics[scale=0.46]{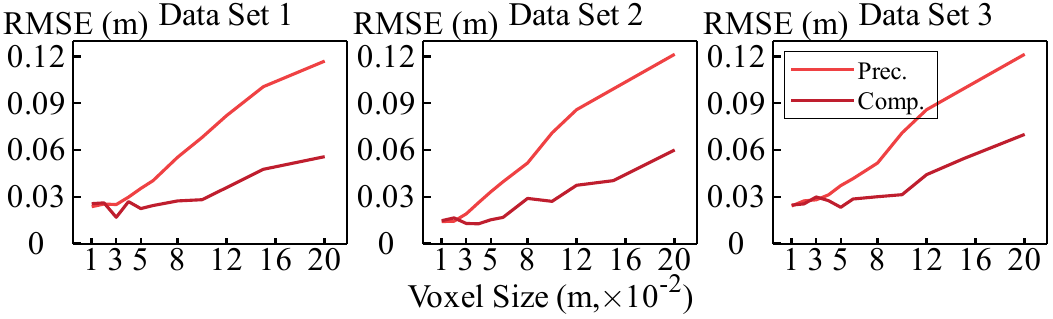}
      \vspace{-0.2cm}
      \caption{The sensitivity analysis of voxel size in filtering. It shows that when the voxel size decreases below a certain threshold, the accuracy of the proposed method no longer improves significantly.}
      \label{figure_sensitive}
      % first line use voxel filtering
      % second line use seg-point with differernt colors 
      % third line use now figure (label the lines gmms and bsplines)
\end{figure}

\begin{figure}[t!]
      \centering
      \includegraphics[scale=0.52]{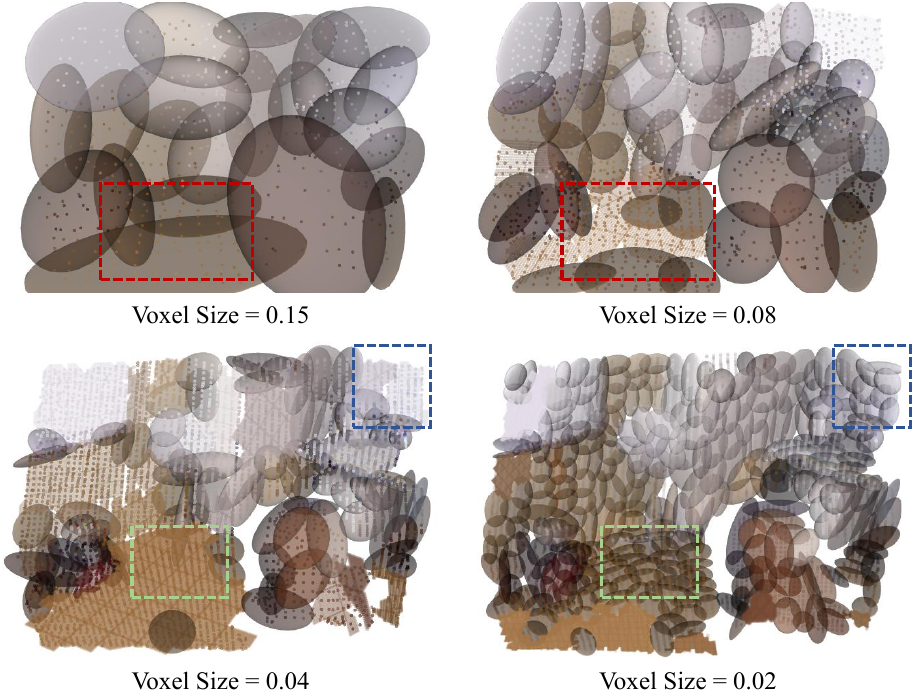}
      \caption{Representations under different voxel sizes. The red box highlights that finer voxel filtering facilitates plane extraction. The green and blue boxes indicate that overly fine voxel filtering amplifies the impact of errors, thereby hindering plane extraction.}
      \label{figure_voxel_size_precision}
      % first line use voxel filtering
      % second line use seg-point with differernt colors 
      % third line use now figure (label the lines gmms and bsplines)
\end{figure}

When adjusting the parameters of the supervoxel filter, achieving a balance between compression ratio and accuracy is difficult, as shown in Fig. \ref{figure_exp_surface_detec}. When the compression ratio is low, the method often produces relatively high errors. If high precision is pursued, this often leads to the excessive use of geometric primitives. Observations from Fig. \ref{figure_overfitting} suggest two main reasons. The position of the voxel may include corners, and supervoxel clustering cannot separate these regions. Corners within such voxels cannot be accurately fitted by Gaussian distributions, planes, or curved surfaces, resulting in high fitting errors. Meanwhile, noise within a voxel may cause incorrect estimation of the normal direction. If the normal of a voxel is not aligned with its neighbors, incomplete merging and disconnected plane detection may occur. By comparison, the proposed method achieves higher accuracy with the same number of primitives. The integrated hierarchical approach of GMM clustering for surface detection is proven to be stable.

In terms of efficiency, our method outperforms the supervoxel approach by factors of \textbf{4.21}, \textbf{8.86} and \textbf{3.78} on the three datasets, respectively. In conclusion, our method can robustly extract the planar and surface features in the point cloud with higher real-time performance.

\subsection{Comparison of Different Geometric Primitives}
\label{sub::compare_model}

In this subsection, plane and surface detection is performed by the proposed method. Several baseline methods are selected for comparison. The B-spline fitting method \cite{b6} performs both boundary fitting and curvature fitting. Its boundary is also modeled with B-spline curves. Gaussian mixture model (GMM) fitting is carried out following \cite{d0}, and quadratic surface fitting follows \cite{a1}.

The RMSE for the quadratic surface resampling cannot be calculated due to the absence of boundary constraints and the existence of multiple solutions. A significant portion of points resampled by quadratic surfaces are located at infinity. All the results are shown in Table \ref{tab:example}.

It is evident that, although the proposed method is slightly less accurate than the B-spline method, its efficiency is significantly improved. Compared with the slightly faster GMM method, the accuracy is doubled. This improvement comes from representing structured point clouds with a more appropriate parametric model, which provides greater practical significance for robotic applications than the accuracy gain alone.

With regard to the reasons for the proposed method's high efficiency, the time complexity of the boundary description is $O(n)$. Furthermore, due to the well-initialized control points, the convex optimization typically converges rapidly.

\subsection{Supplement}
\label{sub::supplement}

\subsubsection{Sensitivity to Voxel Filtering}
In \cite{d0}, the stability of hierarchical Gaussian clustering has been demonstrated. Most merging parameters need little to no manual tuning; only the voxel filter size must be chosen based on the environment. However, the scene size provides only a relative reference and cannot determine the absolute voxel size. As shown in Fig. \ref{figure_sensitive}, we conducted a sensitivity analysis of voxel size with respect to accuracy. The voxel size selected in our experiments ensures that accuracy remains relatively high.

\subsubsection{Upper Limit of Precision}

As illustrated in Fig. \ref{figure_sensitive}, as the voxel size decreases, the accuracy does not improve significantly beyond a certain point. Fig. \ref{figure_voxel_size_precision} shows the corresponding representations. Observation suggests two main reasons for this accuracy limit: (1) fitting errors of the Gaussian model on coarse point clouds, and (2) noise in the point cloud. 
When using GMMs to fit coarse and unstructured point clouds, if the minimum number of points per Gaussian, $N_\text{min}$, is not adjusted, the fitting accuracy of their elliptical boundaries cannot approach zero in the hierarchical structure.
Reducing the voxel size increases the point cloud density, which shrinks the area represented by each Gaussian. This can obscure the shortest-axis feature due to noise, potentially causing failures in plane primitive recognition or normal vector estimation.

\section{Conclusions}

In this paper, we propose a multi-model parametric point cloud representation method capable of robust and real-time detection of curved surfaces. Structured regions in the point cloud are represented using planes and curved surfaces, while highly noisy regions are modeled with Gaussian distributions. Experiments on various datasets show $3.78$ times faster surface detection and $2$ times higher accuracy. Parametric representation of space also improves the efficiency and reduces memory usage in robot mapping and localization. However, these benefits remain limited when only individual frames are processed. In future work, we aim to build upon this foundation to integrate robot pose information, with the goal of constructing a global parametric representation.

% \addtolength{\textheight}{-12cm}  
% This command serves to balance the column lengths on the last page of the document manually. It shortens the textheight of the last page by a suitable amount. This command does not take effect until the next page so it should come on the page before the last. Make sure that you do not shorten the textheight too much.

%%%%%%%%%%%%%%%%%%%%%%%%%%%%%%%%%%%%%%%%%%%%%%%%%%%%%%%%%%%

\end{document}